\documentclass[conference]{IEEEtran}
\IEEEoverridecommandlockouts
\usepackage{cite}
\usepackage{amsmath,amssymb,amsfonts}
\usepackage{algorithmic}
\usepackage{graphicx}
\usepackage{booktabs}

\usepackage{xcolor}
\usepackage{cite}
\usepackage{fancyhdr}

\fancypagestyle{firstpage}{
  \fancyhf{} 
}

\def\BibTeX{{\rm B\kern-.05em{\sc i\kern-.025em b}\kern-.08em
    T\kern-.1667em\lower.7ex\hbox{E}\kern-.125emX}}
\begin{document}
\title{DataAgent: Evaluating Large Language Models' Ability to Answer Zero-Shot, Natural Language Queries
}
\fontsize{12}{12}\selectfont
\author{\IEEEauthorblockN{Manit Mishra}
\IEEEauthorblockA{\textit{Irvington High School} \\
Fremont, United States \\
mshmanit@gmail.com}
\and
\IEEEauthorblockN{Abderrahman Braham}
\IEEEauthorblockA{\textit{}
\textit{Pioneer High School}\\
Sousse, Tunisia \\
br.abderrahman.contact@gmail.com}
\and
\IEEEauthorblockN{Charles Marsom}
\IEEEauthorblockA{\textit{Davis Senior High School}
\\Davis, United States \\
charleshenrymarsom@gmail.com}
\and
\IEEEauthorblockN{Bryan Chung}
\IEEEauthorblockA{\textit{The Loomis Chaffee School} \\
Windsor, United States \\
bryan\_chung@loomis.org}
\and
\IEEEauthorblockN{Gavin Griffin}
\IEEEauthorblockA{\textit{Bellarmine College Preparatory} \\
Sunnyvale, United States \\
gavgriffin563@gmail.com}
\and
\IEEEauthorblockN{Dakshesh Sidnerlikar}
\IEEEauthorblockA{\textit{Rutgers University} \\
New Brunswick, United States\\
dakshesh.sid@gmail.com}
\and 
\IEEEauthorblockN{Chatanya Sarin}
\IEEEauthorblockA{\textit{Bellarmine College Preparatory} \\
Sunnyvale, United States \\
chatanya.sarin@gmail.com}
\and
\IEEEauthorblockN{Arjun Rajaram}
\IEEEauthorblockA{
\textit{University of Maryland, College Park}\\
Frisco, United States \\
arajara1@terpmail.umd.edu}
}

\maketitle

\thispagestyle{firstpage} 

\begin{abstract}
\fontsize{9.5}{9.5}\selectfont
Conventional processes for analyzing datasets and extracting meaningful information are often time-consuming and laborious. Previous work has identified manual, repetitive coding and data collection as major obstacles that hinder data scientists from undertaking more nuanced labor and high-level projects. To combat this, we evaluated OpenAI's GPT-3.5 as a ``Language Data Scientist" (LDS) that can extrapolate key findings, including correlations and basic information, from a given dataset. The model was tested on a diverse set of benchmark datasets to evaluate its performance across multiple standards, including data science code-generation based tasks involving libraries such as NumPy, Pandas, Scikit-Learn, and TensorFlow, and was broadly successful in correctly answering a given data science query related to the benchmark dataset. The LDS used various novel prompt engineering techniques to effectively answer a given question, including Chain-of-Thought reinforcement and SayCan prompt engineering. Our findings demonstrate great potential for leveraging Large Language Models for low-level, zero-shot data analysis.   

\end{abstract}


\section{Introduction}
\fontsize{11.2}{11.2}\selectfont
With a rampant increase in the demand for data processing, specific interest in quickly extrapolating key connections in datasets has grown exponentially. However, the ability of data scientists to meet these demands is waning; though the average number of data scientists within a company has grown from 28 to 50 in the last 9 years \cite{datacareer2022}, and is only expected to continue increasing, that trend cannot compensate for the exponential levels of growth in demand. Put simply, such an increase is unsustainable as interest in processing large amounts of data skyrockets. However, much of the field consists of conducting relatively simple tasks; namely, the act of uncovering patterns and correlations. Data science tasks are often synonymous with repetitive labor, taking much longer than the growing industry desires - analyzing datasets for hidden patterns can be a tedious task, making it a prime candidate for automation through machine learning (ML) techniques. This study aims to examine the efficacy of Large Language Models (LLMs), when paired with an action loop and prompting framework, for analyzing datasets to accomplish various data science tasks.

Significant work already exists in this field. Automatic Prompt Engineers (APEs) have proved useful in demonstrating the power of LLMs to extrapolate correlatory data; when given a set of inputs, APEs are able to identify the most likely ``instruction" for the specific set of inputs \cite{2211.01910}. However, in their current form, APEs have only been identified to work as instruction generators, rather than generators for a roadmap of how to complete a task given an inputted dataset and a human-engineered prompt.  

AutoML models - a general class of interfaces designed to bring ML models to non-ML experts - have also provided another foray into this space \cite{1811.03822}. However, AutoML focuses on building deep learning systems (and other high-level tasks like hyperparameter optimization) without human input (and other high-level tasks like hyperparameter optimization), while our LLM workflow is far more flexible for completing zero-shot data science tasks with specific human instruction. Additionally, a major weakness of AutoML models concerns their prompting; natural-language queries are impractical for models like Auto-PyTorch \cite{2006.13799}. Unlike Large Language Models (LLMs) that are specifically designed to process, understand, and generate human-like text, AutoML models primarily focus on automated machine learning tasks, such as model selection and hyperparameter tuning. For example, Auto-PyTorch can't easily understand a question in plain English, unlike Large Language Models (LLMs). This becomes a prominent issue when users who aren't experts in machine learning need to use these systems. This poses a challenge because AutoML models can't handle conversational language.
\begin{figure}[h]\fontsize{11.2}{11.2}\selectfont
\includegraphics[scale=0.45]{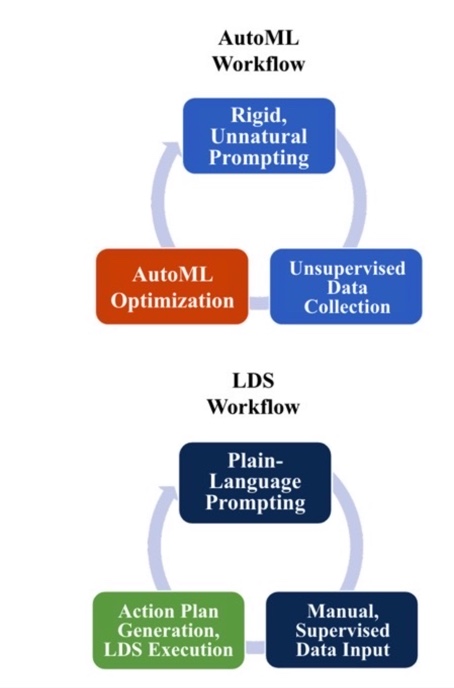}
\caption{Some differences between common AutoML models and the Language Data Scientist.}
\end{figure}

Though existing efforts into improving the accessibility of ML models for non-ML experts are generally well-supported, such efforts are rarely directed towards ameliorating direct, user-generated queries in the field of data science. We have approached this issue using a 3-stage application that heavily integrates GPT language modeling. Upon being prompted with an input dataset and a query related to that dataset, the Language Data Scientist (LDS) will first gather basic background information on the data using Pandas functions, such as pandas.DataFrame.head(), pandas.DataFrame.info(), and pandas.DataFrame.describe(). After acquiring the necessary background information, a GPT model will be prompted to create an ``action plan," which will generate a list of plain-language steps to complete the given task. These steps build on top of each other until finally reaching the final answer to the query. Next, the LDS will use those tasks as a guide to generate lines of code, which will then be run on a low-level executor to complete the task, outputting the answer to the original query. 

To measure the model's accuracy, we generated a set of 15 benchmark datasets along with a corresponding set of manually-created questions and answers for each dataset. Through refinement, reinforcement, and iteration using novel mechanisms - like Chain-of-Thought Prompting and SayCan prompt engineering - the LDS's performance was able to significantly improve\cite{2303.11366} \cite{2201.11903} \cite{2204.01691}. 

The remainder of this paper is organized as follows: Section II provides both a high-level summary and detailed description of the methodology and workflow, in addition to underscoring metrics for measuring the model's accuracy, reinforcement, and efficacy. Section III elucidates overall results, Section IV offers a comprehensive discussion and conclusion of our work, and Section V gives summary acknowledgments. 
\section{Methodology}
\fontsize{11.2}{11.2}\selectfont
\subsection{Summary}\label{A}
Broadly, the methodology for evaluating the LDS's performance consists of three phases: taking in a query and gathering background on a given dataset, formulating a natural-language action plan with the GPT-based action plan generator (AcPG), and systematically feeding the action plan’s steps into the LDS for it to execute, eventually determining a final output. For accuracy measurement, a set of the LDS's predicted outputs for a set of queries is compared to a set of manually calculated and supervised answers for a given benchmark dataset, thus generating an accuracy score.

\subsubsection{Benchmark Datasets \cite{DataAgent2023}}\label{1a}
The LDS is primarily evaluated on 15 benchmark datasets: sets of entries generated using GPT-3.5, ranging from 50 to 300 rows, with different columns consisting of both numerical and categorical data. To best simulate a wide range of data (as would be expected in diverse work environments), the benchmark datasets cover different areas of common data entry, like user phone numbers, names of people, and revenue on advertisements. The datasets were categorized into three size groups: small, medium, and large. Small datasets consist of fewer than 100 rows, while medium datasets range between 100 and 200 rows, and large datasets encompass more than 200 rows. Ground truths for the answers to the questions related to each of the datasets were calculated by hand, using Pandas, NumPy, and various other machine learning libraries.

\subsubsection{Queries}\label{1}
Each benchmark dataset was paired with a set of 15 questions with varying levels of difficulty; an easy question might ask for the number of rows of a Toy Dataset, while a more difficult question might ask about using linear regression to predict a value of the dataset. A sample benchmark dataset is shown in Figure 2.

\begin{figure}[h]\fontsize{10.7}{10.7}\selectfont
\begin{tabular}{lllll} \toprule
\textbf{City} & \textbf{Temp} & \textbf{Humidity} & \textbf{Wind} & \textbf{Clouds} \\ \midrule

New York & 25.3 & 60.0 & 10.9 & PartShade \\
Los Angeles & 30.3 & 50.7 & &  Sun \\
Beijing & 32.7 & 45.2 & 10.0 & PartShade\\
Paris &  & 75.8 & 10.8 & Shade\\ 
Sao Paulo & 35.9 & 60.1 & 12.2 & \\
Moscow & 15.9 & 55.6 & 6.4 & Shade\\
Dubai & 40.5 & 20.7 & 25.3 &  Sun\\
Singapore & 30.5 &  & 15.0 &  PartShade\\
Mumbai & 35 & 70 & 10 & Sun\\
\bottomrule
\end{tabular}

 \caption{A chunk of a sample benchmark dataset, labeled "Cities," containing both numerical and categorical data, along with missing values. The original dataset was medium-sized and had 165 rows.}
\end{figure}

Unless otherwise specified, queries were judged on whether the LDS was able to correctly answer a question, in full, without any additional prompting. Any numerical answers with extremely negligible rounding errors (less than 0.001 percent off from the ground truth) were marked as correct. 

\subsection{Gathering Background Information on a Dataset}\label{BB}
Before an action plan is generated for a given dataset, the LDS gathers basic background information on the inputted data as a part of its context. Our original intention was to feed the entire dataset into the OpenAI API, but this approach was not feasible due to token and query limitations set in place by the API \cite{2303.08774}. As an alternative, when a query is presented for a dataset, the LDS uses GPT-3.5 to generate the code to retrieve any preliminary information needed to answer the question at hand. Along with the background information needed to solve the query, some general information about the dataset using Pandas functions (such as pandas.DataFrame.head() and pandas.DataFrame.info()) are stored as context. This code is then executed and the result is stored in a Python dictionary to be passed as context to the AcPG. The dictionary acts as a way to store the code used to generate the context, along with the context itself.

\subsection{Action Plan Generation}\label{BB}
\subsubsection{Specifics}
The primary portion of the model is a GPT-dependent Action Plan Generator (AcPG), a plain-language roadmap that will be interpreted and fed into the LDS. The AcPG is given basic information about the dataset as well as preliminary information previously generated in past queries, which were all stored in a dictionary. To improve accuracy and help generate action plans, the AcPG utilizes Chain of Thought reasoning \cite{2201.11903}, a technique where a complex task is broken down into a sequence of smaller, explainable steps chained together. Finally, the list of smaller steps is converted into an array to be parsed by the low-level executor.
\subsubsection{SayCan}
The prompts provided to the AcPG are structured based on the SayCan framework \cite{2204.01691}, with distinct sections outlining the situation context, desired response action, declared capabilities, and stipulated needs. While going through the AcPG, GPT-3.5 is instructed to provide code snippets without executing them, focusing solely on preparing the steps required for data analysis. This approach directly addresses the declared capabilities and stipulated needs of our system. As the process unfolds, the context is updated with each step's output, gradually building towards executing the final step that directly answers the initial question. This incremental building of context and action, guided by the SayCan framework, ensures that each step is purposeful and directly contributes to solving the larger query. When compared with traditional natural language prompting, SayCan-formulated prompts assist in drastically improving the AcPG outputs that work with the remaining portion of the LDS. Using SayCan, we structured the answers to suit our purpose of passing the code through the low-level executor.

\begin{figure}[h]\fontsize{10.5}{10.5}\selectfont
\includegraphics[scale=0.6]{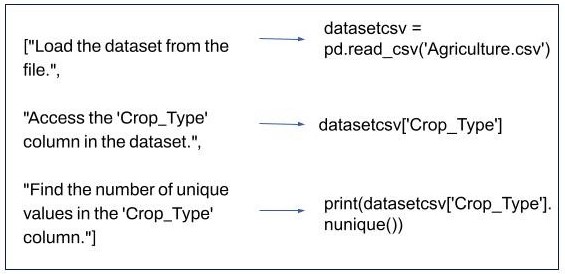}
\fontsize{11}{11}\selectfont
\caption{An example of how natural language steps generated by the AcPG are then translated to code for the executor}
\end{figure}

\subsection{Low-Level Execution}\label{DD}
Once the AcPG generates a sufficient plain-language action plan, along with the specific code from libraries such as NumPy and Pandas necessary to implement that plan, the code will be run on a local low-level executor which calculates a numerical or categorical response to the original query. 
\begin{figure}[h]\fontsize{10.5}{10.5}\selectfont
\includegraphics[scale=0.36]{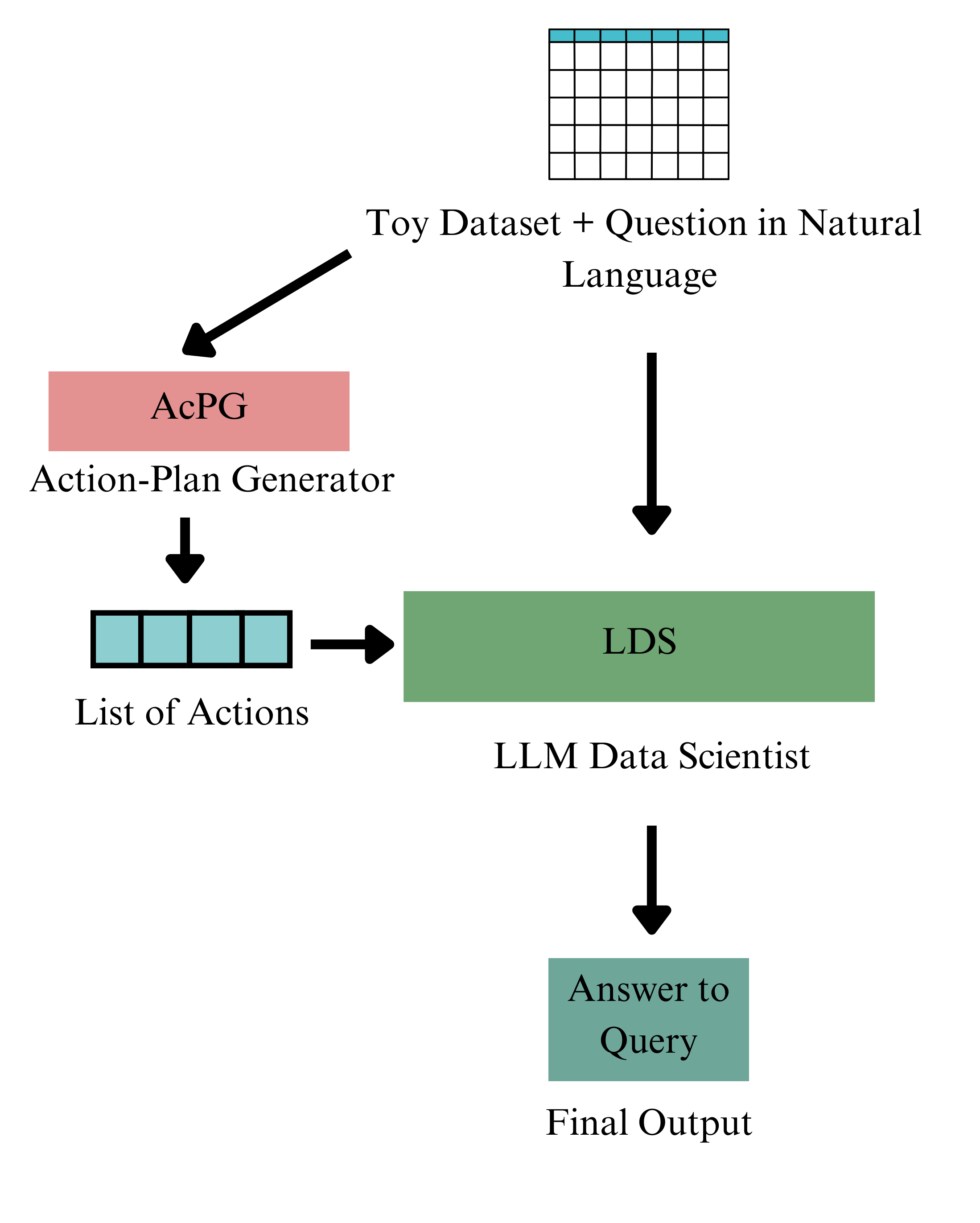}
\fontsize{11}{11}\selectfont
\caption{A broad overview of the model. A natural-language query, paired with an inputted dataset, is sent both to the AcPG and the LDS. The AcPG then generates a plan of action for answering the question with the given data, and an executor in the LDS computes the final output.}
\end{figure}

\subsection{Benchmark Dataset Answer Checker}
In addition to developing the primary model, we also created an answer checker tool to facilitate usage of our benchmark datasets. This tool is designed to assist users in verifying the accuracy of their responses when utilizing our datasets for their own research or practice.

To construct this tool, we first compiled comprehensive dictionaries for each benchmark dataset, cataloging all associated questions and their corresponding correct answers. When a user submits an answer, the tool retrieves the question along with the correct response from the dictionaries. It then employs GPT-4, along with a predefined margin of error, to assess the accuracy of the user's answer. This approach ensures a reliable and efficient means of verification, enabling users to gauge the correctness of their responses effectively.

The incorporation of the GPT-4 API in this process adds an additional layer of sophistication, allowing the tool to handle a range of answer types and nuances. The margin of error parameter provides flexibility, accommodating slight variations in answers that are still fundamentally correct.

\section{Results}
\fontsize{11.2}{11.2}\selectfont
We tested the LDS's ability to accurately extrapolate information within numerical and categorical datasets of three sizes: small, medium, and large, and questions of varying difficulty, language, and style. 

Overall, the LDS answered 74 out of 225 questions correctly, for an accuracy of 32.89 percent. There was a lot of variation in the accuracy rates between Toy Dataset sizes; all individual datasets had accuracies between 20 and 60 percent, and no size or question difficulty level presented an anomaly concerning accuracy. The LDS performed best on Large Benchmark Datasets, answering 36 percent (27 of 75)  questions correctly. A more detailed breakdown of the results is shown in Figure 5.

\begin{figure}[h]\fontsize{11}{11}\selectfont
\begin{tabular}{llll} 
\textbf{} & \textbf{Small} & \textbf{Medium} & \textbf{Large} \\ \midrule

Correct Queries & 25 & 22 & 27\\
Total Queries & 75 & 75 & 75\\
Percent Correct & 33.33 & 29.33 & 36\\

\bottomrule
\end{tabular}

 \caption{A summary of our results, which were generally stable across different Toy Dataset sizes and queries of varying difficulty.}
\end{figure}
As illustrated in Figure 5, minimal variations in model accuracy are observed across various dataset sizes. The LDS consistently demonstrates its performance when exposed to datasets containing varying numbers of rows, spanning from 50 to 300.
\begin{figure}[h]\fontsize{10.5}{10.5}\selectfont
\includegraphics[scale=0.3]{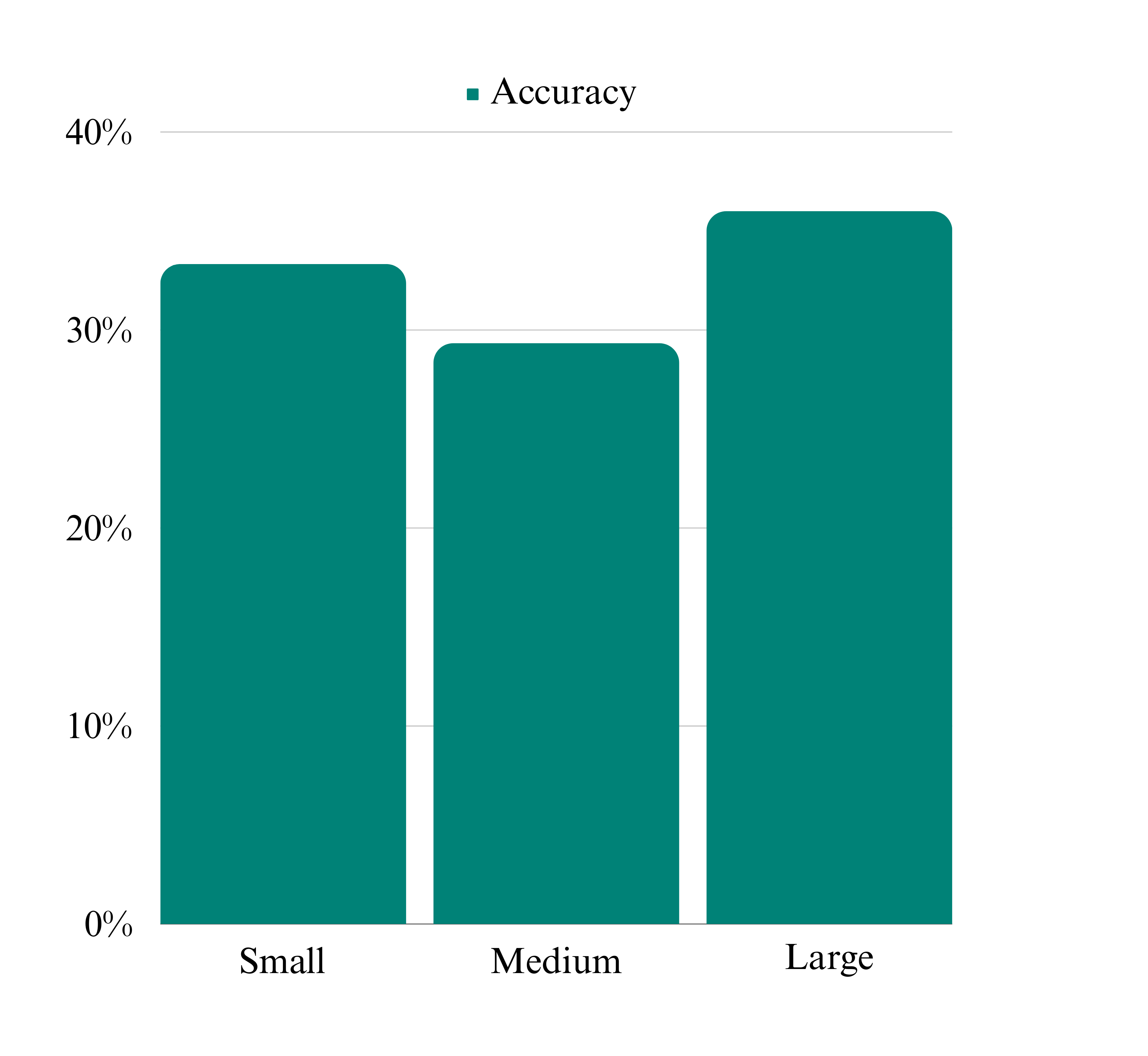}
\fontsize{11}{11}\selectfont
\caption{Model's Accuracy by Dataset Size}
\end{figure}

Out of the 74 incorrect responses, a notable portion was attributed to two primary issues encountered during the query processing. Firstly, there were instances where the GPT model generated incorrect code, involving variables that did not exist in the dataset and calling functions that are not found in Pandas or NumPy. This led to responses that were not only incorrect but also not executable within the framework of our established data analysis environment.

Secondly, we faced challenges with the token limits imposed by the GPT API. The limit of 4096 tokens restricted our ability to provide comprehensive context for larger datasets. This limitation mainly impacted questions that required large amounts of data as context.

\section{Discussion and Conclusion}
\fontsize{11.2}{11.2}\selectfont
To better understand the feasibility of using Large Language Models (LLMs) to find answers for zero-shot queries, specifically in the field of data science, we created and evaluated a sample "LLM Data Scientist" (LDS) that utilized an action plan generator and various methods of reinforcement. Our study demonstrated a large amount of promise in using Large Language Models to accurately perform low-level data analysis, both for numerical and categorical datasets. Below is a brief discussion of notable points of our workflow and its results, along with our takeaways and future plans. 

\subsection{Prompt Rewording}\label{CC}
In any instance where the original question had to be altered to produce correct results, the query was marked as incorrect for accuracy purposes. However, most of these questions produced correct results with minor tweaking; as an example, questions with incorrect answers that asked for ``results" often produced correct responses when ``overall results" was used instead. 

Further work on the benchmark datasets could involve a revision of the questions to enhance their clarity and specificity. By refining the phrasing and structure of the questions, we could potentially reduce ambiguity and improve the model's ability to comprehend and accurately respond to them. 

\subsection{Multiple Answers in One Prompt}\label{CC}
Unexpectedly, the LDS often failed to produce complete answers for queries that required multiple outputs. Most commonly, the model correctly produced one answer to the question but left the rest incomplete. These results were marked as incorrect, but when the questions with multiple parts were rephrased into multiple questions asking for one answer and re-fed into the LDS, the model generally produced correct results. 

\subsection{Edge Cases}
When prompted with questions that didn't fully apply to the context of a dataset, the LDS frequently produced incorrect results as a result of flawed action plan generation. This is consistent with popular findings concerning GPT-model-based generators. Frequently, this manifested in questions that asked for the median value for a categorical (non-numeric) dataset, or in questions that asked for the column with the greatest number of missing values for a dataset that contained no missing values; in this case, the LDS often incorrectly returned the first column in the dataset. 

\subsection{Computing Details and Future Plans}\label{CC}
The relative simplicity of the LDS's structure greatly influenced the types of questions it was able to answer. Though the LDS had shortcomings when given complicated, multi-prompt queries, the model excelled in drawing simple non-obvious connections, data, and conclusions from various Toy Datasets.

The restrictions on the LDS's ability to execute complex queries may also be a result of the GPT model used for the AcPG. We relied on OpenAI's GPT-3.5, a somewhat inferior and outdated model when compared to GPT-4 and other cutting-edge GPT actors. This decision was mainly limited by our budget. 

Considering the vastly improved computing power of novel models when compared to GPT-3.5, we anticipate the AcPG to produce more accurate results when able to harness more capable LLMs; GPT-4 is 82 percent less likely to produce factual errors \cite{openaigpt4}, and includes much greater capabilities for recognizing nuance, context, and complex, multi-part instructions in prompts. Specifically with regard to this increased capability for multi-part prompts, we anticipate that using a GPT-4-based AcPG would allow the LDS to accurately evaluate questions that ask for multiple answers, one of the main drawbacks of the current LDS. 

Additionally, we plan for the AcPG to incorporate refleXion, a  prompt engineering technique that improves model robustness through linguistic feedback rather than weight modification. During training, we will implement a separate refleXion prompt used to generate targeted linguistic adversarial examples that expose biases, oversights, and flaws in the AcPG's reasoning. This prompt will run concurrently with the orignal query. The probing examples given will then be provided as input to the AcPG in successive rounds of generation, pushing it to confront its own limitations as a means of reinforcement. To enable the model to learn from this linguistic feedback over time, we will also an episodic memory buffer that stores examples along with the main model's responses.

Furthermore, while our study has demonstrated the efficacy of Large Language Models (LLMs) for data analysis across a range of dataset sizes, a natural progression for future research would be to explore the performance of these models on even larger datasets. This exploration is crucial as it can reveal how scalable these models are and whether their performance is maintained or even enhanced when dealing with larger volumes of data.

The investigation into larger dataset sizes could potentially unveil unique challenges and opportunities. For instance, larger datasets might introduce more complex patterns or noise, which could test the limits of the model's analytical capabilities. Conversely, they could also provide a richer context for the model to learn from, possibly leading to more accurate or nuanced analyses. Additionally, assessing the performance on larger datasets would provide valuable insights into the computational efficiency and resource utilization of these models, which is a critical factor in real-world applications. 

\section{Acknowledgments}
We would like to thank our mentor, Arjun Rajaram, for his invaluable guidance and support. We also like to thank OpenAI for providing access to their API, which played a crucial role in the development and testing of our workflow.

\fontsize{11.2}{11.2}\selectfont
\section{Abbreviations}
LLM, Large Language Model; LDS, LLM Data Scientist; AcPG, Action Plan Generator; APEs, Automatic Prompt Engineers. 
\bibliographystyle{ieeetr}
\bibliography{ref}

\end{document}